# Decision Support System for Detection and Classification of Skin Cancer using CNN

Rishu Garg[*1[a]], Saumil Maheshwari[*1[b]] and Anupam Shukla[2]

[1[a]] National Institute of Technology, Raipur, India
[1[b]] Atal Bihari Vajpayee- Indian Institute of Information Technology and Management, Gwalior, M.P, India
[2] Indian Institute of Information Technology, Pune, India

[1[a]]rishu.garg.rg06@gmail.com, [1[b]]saumilmaheshwari@yahoo.co.in,

[2]dranupamshukla@gmail.com

* Equal Contributions

**Abstract.** Skin Cancer is one of the most deathful of all the cancers. It is bound to spread to different parts of the body on the off chance that it is not analyzed and treated at the beginning time. It is mostly because of the abnormal growth of skin cells, often develops when the body is exposed to sunlight. The Detection Furthermore, the characterization of skin malignant growth in the beginning time is a costly and challenging procedure. It is classified where it develops and its cell type. High Precision and recall are required for the classification of lesions. The paper aims to use MNIST HAM-10000 dataset containing dermoscopy images. The objective is to propose a system that detects skin cancer and classifies it in different classes by using the Convolution Neural Network. The diagnosing methodology uses Image processing and deep learning model. The dermoscopy image of skin cancer taken, undergone various techniques to remove the noise and picture resolution. The image count is also increased by using various image augmentation techniques. In the end, the Transfer Learning method is used to increase the classification accuracy of the images further. Our CNN model gave a weighted average Precision of 0.88, a weighted Recall average of 0.74, and a weighted f1-score of 0.77. The transfer learning approach applied using ResNet model yielded an accuracy of 90.51%

## 1  Introduction

About more than a million skin cancer cases occurred in 2018 globally [1]. Skin cancer is one of the fastest-growing diseases in the world. Skin cancer occurs mainly due to the exposure of ultraviolet radiation emitted from the Sun. Considering the limited availability of the resources, early detection of skin cancer is highly important. Accurate diagnosis and feasibility of detection are vital in general for skin cancer prevention policy. Skin cancer detection in early phases is a challenge for even the dermatologist.

In recent times, we have witnessed extensive use of deep learning in both supervised and unsupervised learning problems. One of these models is Convolution Neu-ral Networks (CNN) which has outperformed all others for object recognition and object classification tasks. CNNs eliminate the obligation of manually handcrafting features by learning highly discriminative features while being trained end-to-end in a supervised manner.

Convolutional Neural Networks have recently been used for the identification of skin cancer lesions. Several CNN models have successfully outperformed trained human professionals in classifying skin cancers. Several methods like transfer learning using large datasets have further improved the accuracy of these models.

VGG-16 is a convolutional neural system that is prepared on more than a million pictures from the ImageNet database. The system is 16 layers profound and can arrange pictures into 1000 item classifications, for example, console, mouse, pencil, and numerous creatures. Accordingly, the system has learned rich component portrayals for a wide scope of pictures. The system has a picture info size of 224-by-224. The model accomplishes 92.7% top-5 test precision in ImageNet, which is a dataset of more than 14 million pictures having a place with 1000 classes.

In this paper, we have generated a CNN model that analyses the skin pigment lesions and classifies them in using a publicly available dataset by employing various deep learning techniques. We have improved the accuracy of classification by implementing CNN and transfer learning models. We have tested our model using the publicly available HAM-10000 dataset

## 2  Literature Review

CNNs have been used frequently in the field of medical image processing image classification and so on [2]. CNNs have already shown inspiring outcomes in the domain of microscopic images classification, such as: human epithelial 2 cell image classification [3], diabetic retinopathy fundus image classification [4], cervical cell classification [5] and skin cancer detection [6-9].

Brinker et al. [10] presented the first systematic study on classifying the skin lesion diseases. The authors specifically focus on the application of CNN for the classification of skin cancer. The research also discusses the challenges needed to be addressed for the classification task. Han et al. in [11] proposed a classifier for 12 different skin diseases based on clinical images. They developed a ResNet model that was fine-tuned with 19,398 training images from Asan dataset, MED-NODE dataset, and atlas site images. This research does not include the patients with different set of ages.

Authors in [12] presented the first comparison of CNN with the international group of 58 dermatologist for the classification of the skin cancer. Most dermatologists were outperformed by the CNN. Authors concluded that, irrespective of any physicians' experience, they may benefit from assistance by a CNN's image classification. Google's Inception v4 CNN architecture was trained and validated using dermoscopic images and corresponding diagnoses. Marchetti et al. [13] performed cross-sectional study using 100 randomly selected dermoscopic images (50 melanomas, 44 nevi, and 6 lentigines). This study was performed over an international computer vision melanoma challenge dataset (n = 379). Authors developed a fusion of 5 methods for the classification purpose. In [14] authors trained a CNN-based classification model on 7895 dermoscopic and 5829 close-up images of lesions. These images were excised at a primary skin cancer clinic between January 1, 2008, and July 13, 2017. Further, the testing of the model was done on a set of 2072 unknown cases and compared with results from 95 human raters who were medical personnel.
Mostly existing research considers binary classification, whether the cancer is melanomous or not and small work is performed for classification of general images. but their result is not very optimal. The existing algorithms used for the detection and classification of skin cancer disease uses machine learning and neural network algorithms.

## 3  Methodology

This Section will emphasis over the methodology adopted for the classification task. Over all steps of the methodology is shown in figure 1.

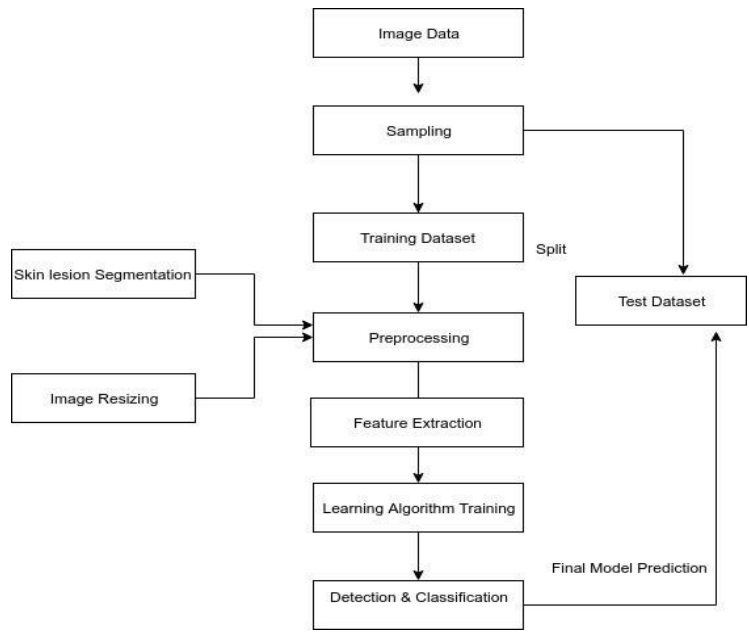

**Fig. 1.** The Flowchart of the methodology implemented.

### 3.1  Dataset Description

We used the MNIST HAM-10000 dataset for Skin Cancer which is available on kaggle [15,16]. It contains 10015 images of skin pigments which are divided amongst seven classes. The number of images present in the dataset is enough to be used for different tasks including image retrieval, segmentation, feature extraction, deep learning, and transfer learning, etc

## 3.2 Preprocessing

The dataset had to be cleaned and organized before being fed into the model. However, the data is highly imbalanced with the lesion type 'melanocytic nevi' comprising of more than fifty percent of the total dataset. We have applied several pre-processing networks to enhance the learnability of the network. We have performed Data Augmentation to avoid the overfitting of data. We have created several copies of the existing dataset by translating, rotating and zooming the images by various factors. Also, in this paper, we have increased the contrast of skin lesions using Histogram Equalization.

# 4 Method

Convolution Neural Networks and Transfer Learning methods are used for the classification task. For the purpose of Transfer learning, deep learning models pre-trained on ImageNet dataset was used. It consists of a little more than 14 million labeled images belonging to more than 20,000 classes. Later on, these pre-trained models are further trained on the HAM10000 dataset by adding some additional layers to the models and freezing some of the initial layers. We also applied different learning algorithms like XGBoost, SVM and Random Forest Algorithms to perform the classification task in the HAM10000 dataset to compare the reuslts

## 4.1 Convolution Neural Network

Convolutional neural networks were inspired by biological processes. In these, the connectivity pattern between neurons of a network resembles the organization of the animal visual cortex. The response of an individual cortical neuron in a restricted region of the visual field is known as the receptive field. The receptive fields of different neurons partially overlap such that they cover the entire visual field. CNN is comprised of three stages of neural layers to assemble its structures: Convolutional, Pooling, and Fully-Connected.

**Convolution Layer:** In CNN the main layer is convolutional layer. In this layer the result of the output layer is gotten from the input by filtering in specific condition. This layer is constructed by the neurons which is in the shape of cubical blocks.

**Max-pooling layer:** Pooling layer executes the next operation after each convolution layer. These layers are utilized to minimize the size of the neurons. These are small rectangular grids that acquires small portion of convolutional layer and filters it to give a result from that block. The most commonly used layer is max pooling that fetch that maximum pixel from the block.

**Fully Connected layers:** The final layer of a convolutional neural network(CNN) is a fully connected layer that is formed from the attachment of all preceding neurons. It reduces the spatial information as it is fully connected like in artificial neural network. It contains neurons beginning at input neurons to the output neurons.

### 4.2   Transfer Learning

Transfer learning is a technique in which a pre-trained model is used on another dataset. This technique is mainly used when there are not enough input data to properly train the model. In such cases, a different model, which is already trained in a different large dataset is used. Here, we used some of the models which were pre-trained in the ImageNet dataset that contain millions of images that are associated with 1000 classes. These models are further appended with different untrained layers and further trained in the HAM10000 dataset. The architecture of the models used namely VGG16 is shown in figure 2

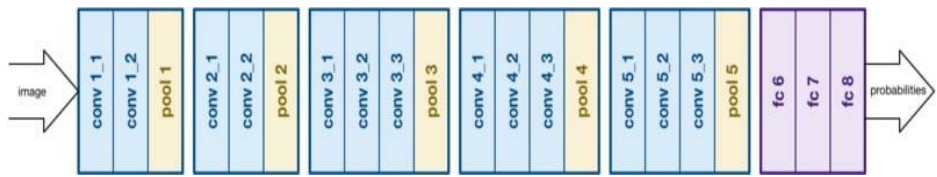

**Fig. 2.** Architecture of VGG16.

## 5   Implementation

In this research, we trained a CNN network to classify the given images in the MNIST dataset in their respective classes. Architecture of CNN model used for the classification is shown in table 1. Python programming language is used for programming of the model. Keras is the primary deep learning library used for computation purposes along with numpy. Matplotlib was used for graph plotting purposes. For fair splitting of data, we have incorporated stratified splitting of data for training and testing purposes. For further tackling the class imbalance, we have incorporated the concept of class weights in which the misclassification of a minority class is penalized heavily. Dropout was also used as regularization technique to ensure better generalization of model over test data. Transfer learning technique was also used to compare the accuracy of the model with that of the proposed deep learning model. This model is then trained with the new dataset by freezing more than 70% of the layers in the VGG network. The last few layers are re-trained with the new dataset to mould the network according to the new dataset. Finally, this network is concluded by adding a few fully connected layers.

We used the Transfer Learning methods using models like ResNet or VGG16 which was pre-trained with the ImageNet dataset. Adam optimizer was used for optimization purpose and categorical cross-entropy loss was used for calculating the loss in the model.

We have also tested the dataset on various Machine Learning Algorithms including Random Forest, XGBoost and Support Vector Machines.

**Table 1.** Architecture of CNN Implemented.

| Layers | Output Size | Kernel size |
|---|---|---|
| Input | 3 x 512 x 512 | |
| Convolution | 32 x 510 x 510 | 3 x 3 |
| ReLu Activation | 32 x 510 x 510 | |
| Convolution | 32 x 508 x 508 | 3 x 3 |
| ReLu Activation | 32 x 252 x 252 | |
| Convolution | 32 x 506 x 506 | 3 x 3 |
| ReLu Activation | 32 x 506 x 506 | |
| Max Pooling | 32 x 253 x 253 | |
| Convolution | 64 x 251 x 251 | 3 x 3 |
| ReLu Activation | 64 x 251 x 251 | |
| Convolution | 256 x 57 x 57 | 3 x 3 |
| ReLu Activation | 256 x 57 x 57 | |
| Convolution | 256 x 53 x 53 | 5 x 5 |
| ReLu Activation | 256 x 53 x 53 | |
| Global Pooling | 256 | |
| Dense | 4096 | |
| ReLu Activation | 4096 | |
| Dense 2 | 5 | |
| Sigmoid Activation | 5 | |

## 6  Results

The models created were trained by using the balanced and resized images from the dataset. Kaggle's kernels were used for performing the training testing and validation of the models. The number of epochs for which the models were trained was 50. We then calculated the confusion matrix, shown in Figure 3 and evaluated the models using the overall accuracy of classification. We have developed a CNN and model for the classification of skin cancer. Our CNN model gave a weighted average value of Precision of 0.88, a weighted Recall average of 0.74 and a weighted f1-score of 0.77. Table 2 shows the result matrix of different Transfer Learning Models. Table 3 shows the performance of our CNN model on the various classification of various classes. We also tried to test our dataset of different models like Random Forest, XGBoost and Support Vector Classifiers. However, we did not see promising results in these learning algorithms. The results are displayed in Table-5

**Table 2.** Results Summary of Models.

| Sr. No | Model Name | Accuracy |
|---|---|---|
| 1 | ResNet | 90.5 % |
| 2 | VGG16 | 78 % |

**Table 3.** Results Summary of self-made models

| Sr. No | Class of Lesion | Precision | Recall | f1-score |
|---|---|---|---|---|
| 1 | Actinic Keratoses | 0.27 | 0.62 | 0.37 |
| 2 | Basal cell carcinoma | 0.45 | 0.73 | 0.56 |
| 3 | Benign keratosis | **1.00** | 0.09 | 0.17 |
| 4 | Dermatofibroma | 0.08 | 0.67 | 0.14 |
| 5 | Melanoma | 0.21 | 0.59 | 0.30 |
| 6 | Melanocytic nevi | **0.95** | **0.82** | **0.88** |
| 7 | Vascular skin lesions | 0.67 | 0.73 | 0.70 |

**Table 4.** Results Summary of self-made CNN (average value)

| Sr. No | Entity | Precision | Recall | f1-score |
|---|---|---|---|---|
| 1 | Micro Average | 0.74 | 0.74 | 0.74 |
| 2 | **Weighted Average** | **0.88** | **0.74** | **0.77** |

**Table 5.** Result Summary of other Machine Learning models

| Sr. No | Model | Accuracy |
|---|---|---|
| 1. | Random Forest | 65.9% |
| 2. | XGBoost | 65.15% |
| 3. | Support Vector Classifier | 65.86% |

**Fig. 3.** Confusion Matrix of ResNet model.

## 7  Conclusion

The proposed method follows an approach in which first step is feature extraction and then these features are used to train and test the transfer learning model. Based on the observation, we have concluded that the Transfer Learning mechanism can be applied to HAM10000 dataset to increase the classification accuracy of skin cancer Lesions. Also, we have found that the ResNet model which is pre-trained in ImageNet Dataset can be very helpful is the successful classification of cancer lesions in the HAM1000 Dataset. We have further seen that Learning algorithms like Random Forest, XGBoost and SVMs are not very effective for classification tasks in the HAM10000 dataset. Encouraged by these outcomes, future work will include the improvement of prediction result and classification accuracy.